\documentclass[10pt,journal,compsoc]{IEEEtran}
\usepackage{multirow}
\usepackage{amsfonts}
\usepackage{rotating}
\usepackage{balance}
\usepackage{enumitem}
\usepackage{bbding}

\ifCLASSOPTIONcompsoc
  \usepackage{cite}
\else
  \usepackage{cite}
\fi
\ifCLASSINFOpdf
\else
\fi

\usepackage{xcolor}
\usepackage{graphicx}
\usepackage{subcaption,siunitx,booktabs,caption}

\newcommand{\respond}[1]{\textcolor{black}{#1}}

\begin{document}

\title{Edge-Cloud Polarization and Collaboration: A Comprehensive Survey for AI}

\author{Jiangchao~Yao, Shengyu~Zhang, Yang~Yao, Feng~Wang, Jianxin~Ma, Jianwei~Zhang, Yunfei~Chu, Luo~Ji, Kunyang~Jia, Tao~Shen, Anpeng~Wu, Fengda~Zhang, Ziqi~Tan, Kun~Kuang, Chao~Wu, Fei~Wu\Envelope,
~\IEEEmembership{Senior Member,~IEEE,}   
Jingren~Zhou,
~\IEEEmembership{Fellow,~IEEE,} 
Hongxia~Yang\Envelope 
~\IEEEmembership{Member,~IEEE,}
\IEEEcompsocitemizethanks{\IEEEcompsocthanksitem 
J. Yao, Y. Yao, F. Wang, J. Ma, J. Zhang, Y. Chu, L. Ji, K. Jia, J. Zhou and H. Yang are with DAMA Academy, Alibaba Group. \\
E-mail: $\{$jiangchao.yjc, yy222244, wf135777, jason.mjx, zhangjianwei.zjw, fay.cyf, jiluo.lj, kunyang.jky, jingren.zhou, yang.yhx$\}$@alibaba-inc.com
\IEEEcompsocthanksitem S. Zhang, T. Shen, A. Wu, F. Zhang, Z. Tan, K. Kuang and C. Wu are with Zhejiang University. F. Wu is with Institute of Artificial Intelligence, Zhejiang University, Hangzhou, China and Shanghai Institute for Advanced Study, Zhejiang University, Shanghai, China\\
E-email: $\{$sy\_zhang, tao.shen, anpwu, fdzhang, tanziqi, kunkuang, chao.wu, wufei$\}$@zju.edu.cn
\IEEEcompsocthanksitem Wu, F. and Yang, H. are corresponding authors.
}
}

\markboth{Journal of \LaTeX\ Class Files,~Vol.~14, No.~8, August~2015}%
{Shell \MakeLowercase{\textit{et al.}}: Bare Demo of IEEEtran.cls for Computer Society Journals}
%



\IEEEtitleabstractindextext{%
\begin{abstract}
Influenced by the great success of deep learning via cloud computing and the rapid development of edge chips, research in artificial intelligence (AI) has shifted to both of the computing paradigms, i.e., cloud computing and edge computing. In recent years, we have witnessed significant progress in developing more advanced AI models on cloud servers that surpass traditional deep learning models owing to model innovations (e.g., Transformers, Pretrained families), explosion of training data and soaring computing capabilities. However, edge computing, especially edge and cloud collaborative computing, are still in its infancy to announce their success due to the resource-constrained IoT scenarios with very limited algorithms deployed. In this survey, we conduct a systematic review for both cloud and edge AI. Specifically, we are the first to set up the collaborative learning mechanism for cloud and edge modeling with a thorough review of the architectures that enable such mechanism. We also discuss potentials and practical experiences of some on-going advanced edge AI topics including pretraining models, graph neural networks and reinforcement learning. 
Finally, we discuss the promising directions and challenges in this field.  
\end{abstract}

\begin{IEEEkeywords}
Cloud AI, Edge AI, Edge-Cloud Collaboration, Hardware.
\end{IEEEkeywords}}

\maketitle

\IEEEdisplaynontitleabstractindextext

%
\IEEEpeerreviewmaketitle

\IEEEraisesectionheading{\section{Introduction}\label{sec:introduction}}
Cloud computing concerns the provisioning of adequate resources to construct a \respond{cost-efficient computing paradigm for numerous applications~\cite{duc2019machine} while edge computing can offer low-latency services}. For cloud computing, it has flourished for a long period in the past decades and achieved a great success in the market, \textit{e.g.,} Amazon EC2, Google Cloud and Microsoft Azure. According to the recent analysis~\cite{size2020share}, the global cloud computing market size was valued at USD 274.79 billion in 2020 and is expected to grow at a compound annual growth rate of 19.1\% from 2021 to 2028. Concomitantly, Artificial Intelligence (AI), especially the compute-intensive Deep Learning~\cite{lecun2015deep}, has enjoyed the tremendous development with the explosion of cloud computing. Nevertheless, the rapid increase of Internet of Things (IoT)~\cite{ashton2009internet} raises an inevitable issue of the data transfer from the edges to the data centers in an unprecedented volume. Specifically, about 850 ZB data is generated by IoT at the network edge by 2021, but the traffic from worldwide data centers only reaches 20.6 ZB~\cite{Cisco:2021}. This drives the emergence of 
the decentralized computing paradigm, edge computing, which turns out to be an efficient and well-recognised solution to reduce \respond{the transmission delay}. Similarly in algorithmic application layer of edge computing, there is an urgent need to push the AI frontiers to the edges so as to fully unleash the potential of the modeling benefits~\cite{zhou2019edge}. 

The existing two computing paradigms, cloud computing and edge computing, polarize the algorithms of AI into different directions to fit their physical characteristics. For the former, the corresponding algorithms mainly focus on the model performance in generalization~\cite{bousquet2003introduction}, robustness~\cite{hansen2011robustness}, fairness~\cite{barocas2017fairness} and generation~\cite{gregor2015draw,fedus2018maskgan} \textit{etc.}, spanning from computer vision (CV), natural language process (NLP) to other industrial applications. To achieve better performances, a large amount of research from the perspectives of the data, the model, the loss and the optimizer is devoted to exploring the limit under the assumption of sufficient computing power and storage. For example, the impressive Generative Pre-trained Transformer 3 (GPT-3) that has 175 billion parameters and is trained on the hundred-billion scale of data, can produce human-like texts. The AlphaFold~\cite{jumper2021highly} with the elaborate network design for the amino acid sequence is trained with a hundred of GPUs and makes the astounding breakthrough in highly accurate protein structure prediction. Nowadays, cloud computing is continuously advancing AI widespread to various scientific disciplines and impact our daily lives.

\begin{figure*}[ht!] 
\begin{center}
\includegraphics[width=0.9\linewidth]{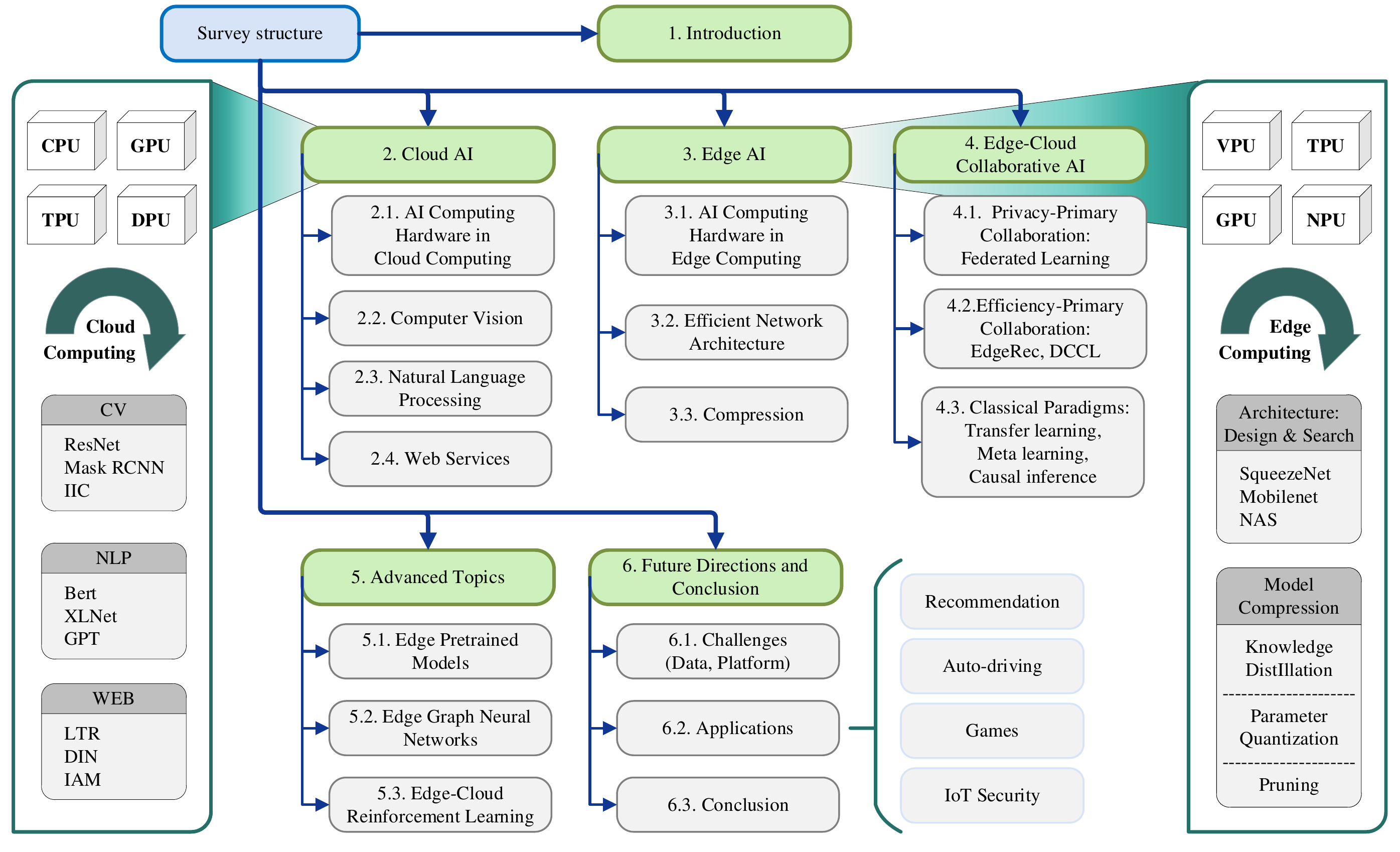}
\caption{Layout of the Survey.}\label{fig:overview}
\end{center} 
\vspace{-0.2in}
\end{figure*} 

However, in terms of edge computing, it is still in its infancy to announce the success due to the constrained IoT resources. There are several critical limitations to design AI algorithms that run on IoT devices while maintaining the model accuracies. The most critical factor is the \emph{processing speed} for the applicability of any edge application~\cite{Nielsen:1993}. We usually use throughput and latency for the measurement, which respectively count the processing rate and characterize the time interval between a single input and its response. Second, \emph{memory} like RAM and cache is the critical resource for building AI edge applications. Machine learning algorithms often take a significant portion of memory during model building with the storage of model parameters and other auxiliary variables. Except the storage, querying the model parameters for processing and inference is both time-consuming and energy-intensive~\cite{sze2017efficient}, yielding \emph{power consumption} and \emph{thermal performance} both are bottlenecks.

Given the above constraints, edge computing polarizes AI to a new research era, namely \emph{edge intelligence} or \emph{edge AI}\footnote{Edge computing originally refers to the computing in edge infrastructures like rounters. We here extend it to a broader range including smart phones \textit{etc}. Edge AI therein we talk also includes on-device AI.}\cite{Wang:2018, Li:2018,jeronimo2017mobile}. Contrary to entirely relying on the cloud, edge AI makes use of the edge resources to gain further AI insights. For example, one direction is to make the model lightweight. One representative work is MobileNet~\cite{howard2017mobilenets}, which reduces the number of parameters from 29.3 million to 4.2 million and the number of computations by a factor of 8 while only losing 1\% accuracy. Furthermore, EfficientNet~\cite{Tan:2019} is proposed to systematically scale up CNN models in terms of network depth, width and input image resolution. It achieved the state-of-the-art performance via 8 times lesser memory but 6 times faster inference.
Major enterprises, such as Google, Microsoft, Intel, IBM, Alibaba and Huawei, have put forth pilot projects to demonstrate the effectiveness of edge AI. These applications cover a wide spectrum of AI tasks, \textit{e.g.,} live video analytics~\cite{Ananthanarayanan:2017}, cognitive assistance for agriculture~\cite{Ha:2014}, smart home~\cite{Jie:2017} and the industrial IoT~\cite{Li:2018}.

Despite distinct characteristics of cloud computing and edge computing, a complete real-world system usually involves their collaboration from the physical aspects to the algorithmic aspects. We term this cooperation as \textit{Edge-Cloud collaboration}, and actually some early explorations have been made in both academia and industry, \textit{e.g.,} federated Learning (FL)~\cite{mcmahan2017communication}. Recently, a successful industrial practice is the Taobao EdgeRec System~\cite{edgerec}. In EdgeRec, the memory-consuming embedding matrices are deployed on the cloud and the lightweight component executes the real-time inference on the edge. Similarly in~\cite{ding2020cloud}, a CloudCNN provides the soft supervision to each local EdgeCNN for the edge training and simultaneously, the EdgeCNN performs the real-time inference interacting with the input.

However, the huge diversity in different algorithmic areas prevents a systematic review from the complete edge-cloud scope. For example, the transformer~\cite{devlin2019bert,khan2021transformers} conquers various benchmark completions in CV and  NLP, but we did not witness its much progress on edge. Edge computing~\cite{zhou2019edge} has the quite promising IoT applications in the future intelligent life, but without the auxiliary from cloud computing, the powerful large models will be extravagant to most of us. Although FL~\cite{kairouz2019advances} is elegant to protect the user privacy, given the heterogeneity and hardware constraints, we still have to explore some new solutions to remedy the performance degradation. All these concerns might be trapped in the local area and some works in the cross-field might enlighten us to find a better way. Motivated by this intuition, we conduct a comprehensive and concrete survey about the edge-cloud polarization and collaboration. Specifically, we are the first to set up the collaborative learning mechanism for cloud and edge modeling. To summarize, we organize the survey of this paper as follows and in Fig.~\ref{fig:overview}: 
\begin{itemize}[leftmargin=*]
    \item Section 2 gives an overview of cloud AI; 
    \item Section 3 discusses edge AI;
    \item Section 4 reviews the architectures that enable the collaboration between cloud and edge AI; 
    \item Section 5 discusses some on-going advanced topics, including pretraining models (PTM), graph neural networks (GNN) and reinforcement learning (RL) models that are deployed on edge; 
    \item We conclude the paper in Section 6 with some possible future directions and challenges. 
\end{itemize}

\section{Cloud AI}
AI has achieved the tremendous development in the recent years, even outperformed the human performance in a range of open-source competition benchmarks~\cite{silver2017mastering,grace2018will}. One main reason behind this success is the benefit from the cloud computing, \textit{i.e.,} the large-scale distributed cluster, which greatly accelerate the training of AI models. We term them as \emph{cloud AI}. Considering the real-world AI applications depend on the information carriers such as image and text, we review some computing hardwares and cloud models in the areas of CV, NLP and web services as exemplars.

\begin{figure*}[ht!] 
\begin{center}
\includegraphics[width=\linewidth]{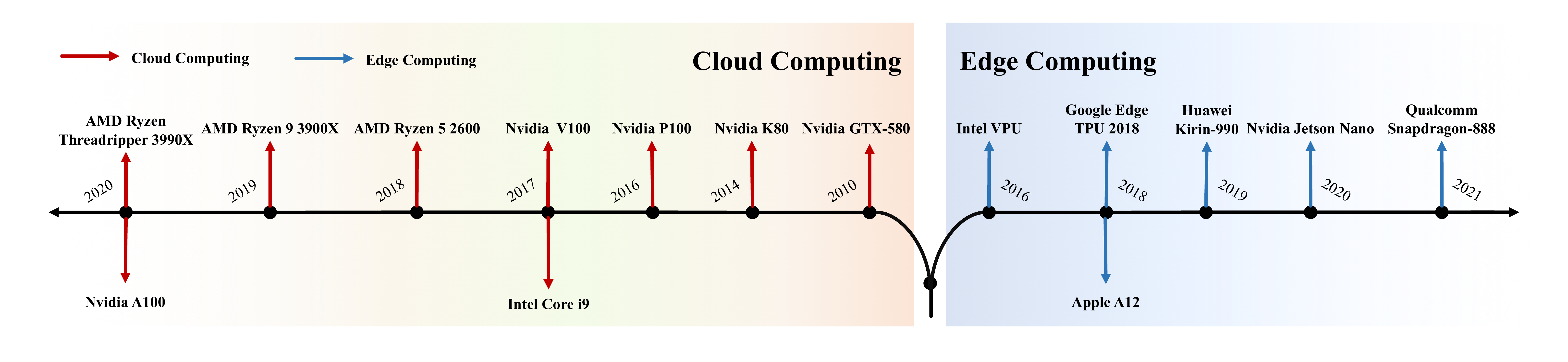}
\caption{\respond{Some exemplar computing hardwares widely used by the algorithmic development of AI models in cloud computing and edge computing, collected according to the open websites as shown in the paragraphs without any business purposes.} }\label{fig:hardware}
\end{center} 
\vspace{-0.2in}
\end{figure*}

\begin{table}[t!]
    \centering
    \begin{tabular}{c|c}
        \toprule
        acronym & description \\
        \midrule
        IoT & Internet of Things \\
        CV & Computer Vision \\
        NLP & Natural Language Processing \\
        CNN & Convolutional Neural Network \\
        RNN & Recurrent Neural Network \\
        DNN & Deep Neural Network \\
        GNN & Graph Neural Networks \\
        RL & Reinforcement Learning \\
        PTM & Pretraining Modeling \\
        FL & Federated Learning \\
        FRL & Federated Reinforcement Learning \\
        KD & Knowledge Distillation \\
        QA & Question Answering \\
        NAS & Neural Architecture Search \\
        PQ & Product Quantization \\
        \bottomrule
    \end{tabular}
    \caption{\respond{Description of popular acronyms in our survey.}}
    \label{tab:my_label}
\end{table}

\subsection{\respond{AI Computing Hardware in Cloud Computing}}
\respond{There are many factors on hardware and software as well as the customer-friendly resource management technology (\textit{e.g.,} the virtualization technology~\cite{lombardi2011secure,jain2013network,tang2021slicing}) to thrive on the great success of cloud computing. We kindly refer readers to~\cite{duc2019machine} for a complete analysis. In this part, we specially recap some computing hardware from the mainstream companies, which plays the crucial roles in the AI flourish.}

\subsubsection{CPU}
\respond{The high performance of Central Processing Unit (CPU) is important to accelerate the computing-intensive AI tasks. In the last decade, there are many excellent breakthroughs in the processor design. As one exemplar in Intel, the Core i9-7920X CPU is the fastest processor for the general computation in the Core X-series processors.} The competitor AMD released the Ryzen-series of products like Ryzen 5 2600 and Ryzen 9 3900X, both for desktops and IoTs based on the Zen micro-architecture. Recently, the AMD Ryzen Threadripper 3990X that has the best performance is the flagship of Ryzen series especially designed for the edge machine learning. \respond{We enumerate their release time of above CPU processors in Fig.~\ref{fig:hardware} according to the time-varying website ranking\footnote{https://benchmarks.ul.com/compare/best-cpus} of Intel and AMD, but without any business purposes. Note that, the general CPU server is slow for the large-scale training of DNNs, and thus some specific speedup technologies from the task scheduling are usually deployed~\cite{awan2017depth} or mixed with the following modules~\cite{jiang2020unified,jain2021optimizing}.}

\subsubsection{GPU}
Graphics Processing Unit (GPU) was originally designed to create images for computer graphics and video games. In early 2010, researchers found that GPUs can be used to accelerate calculations of DNNs, based on which AlexNet achieved a great success in the ImageNet challenge~\cite{krizhevsky2017imagenet}. The GPU behind AlexNet is Nvidia GeForce GTX 580, which is a high-end graphics card launched by NVIDIA in 2010. Subsequently, NVIDIA released the Tesla K80 and Tesla P100 that further improved the computation performance. Recently, Tesla V100 as well as Tesla A100 again catches the eyes due to the impressive PTM~\cite{floridi2020gpt,m6,m6-t} in language understanding and image generation tasks. \respond{We similarly include the release time of some popular GPUs in Fig.~\ref{fig:hardware} according to the time-varying open ranking\footnote{https://analyticsindiamag.com/top-10-gpus-for-deep-learning-in-2021/}.}

\subsubsection{TPU and DPU}
Tensor Processing Unit (TPU) is a dedicated integrated circuit developed and used in Google Cloud, \respond{which is superior in the DNN computation. Its first generation is identified by the high-bandwidth loop~\cite{Norrie_Patil_Yoon_Kurian_Li_Laudon_Young_Jouppi_Patterson_2021} and the second generation is known by faster cost-effective mixed precision training~\cite{Wang_Wei_Brooks_2019}}. The third generation provides significant performance benefits over TPUv2 and fits better in the larger-scale network architecture. In Google I/O 2021, Google launches the latest TPUv4 with nearly two times performance over TPUv3. \respond{Data Processsing Unit (DPU) is a recently proposed programmable specialized electronic circuit for data-centric computing, which frees the computing load of CPU or GPU in the hardware aspect\footnote{https://blogs.nvidia.com/blog/2020/05/20/whats-a-dpu-data-processing-unit/}. It is practically useful for the acceleration of AI models in face of the expensive loading and offloading cost of large-scale neural networks like PTM.}

\subsection{Computer Vision}
Among various research areas, CV is a longstanding field, which asks computers to derive the semantics from digital images, videos, and other visual inputs. In this part, we review some classical tasks, \textit{i.e.,}, Image Recognition, Object Detection and Image Segmentation.

Image recognition involves analyzing images and identifying objects, actions, and other elements in order to draw conclusions. In the 1880s, the predecessor of CNN ``neocognitron" was proposed. Then, Lecun \textit{et al.,}~\cite{3lecun1989backpropagation} proposed the CNN operator and successfully applied it to handwritten digital character recognition, with an error rate of less than 1\%. In 2012, AlexNet~\cite{krizhevsky2017imagenet} won the ImageNet competition \cite{deng2009imagenet} and incorporated many enticing techniques such as ReLU nonlinearity, dropout, and data augmentation. Following AlexNet, deeper structures like VGG-16~\cite{17simonyan2014very} and GoogleNet~\cite{18szegedy2015going} were explored. To better handle gradient vanishing issue, ResNet~\cite{he2016deep} used a skip-connection scheme to form a residual learning framework. As such, it scales convolutional layers up to 152 and \respond{becomes} the winner of 
multiple vision tasks in ILSVRC 2015 and MS COCO 2015.

Object detection identifies and locates objects in an image/scene.
In recent years, one-stage and two-stage object detectors~\cite{20law2018cornernet,21lin2017focal,22redmon2016you,23cai2018cascade} have achieved noticeable improvements. However, these methods rely on deep convolution operations to learn intensive features, resulting in a sharp increase in the cost of computing resources and an apparent decrease in detection speed~\cite{6guo2021distilling}. Therefore, how to address these problems and enable real-time detection becomes an important line of research in object detection. Recently, pruning techniques and knowledge distillation are effective tools to build a lightweight object detector~\cite{28chen2017learning,29wang2019distilling}.

Image segmentation~\cite{9cho2021picie} is another fascinating technique that separates an image into several recognizable segments. Apparently, image segmentation requires tremendous human effort in labeling pixels. Recently, IIC~\cite{34ji2019invariant} uses the mutual information-based clustering to output a semantic segmentation probability map in image pixels. AC identifies
probabilities of pixels over categories by an autoregressive model and maximizes mutual information across two different ``orderings"~\cite{35ouali2020autoregressive,36oord2016conditional}. In addition to generic domains, many experts also focus on investigating the domain-specific image segmentation techniques ~\cite{37jha2020doubleu,38ronneberger2015u,39zhou2019unet++}.

Except above subareas, great success also has been made in other CV tasks such as super-resolution~\cite{12song2021addersr}, image restoration~\cite{45pan2021exploiting}, and image generation~\cite{ramesh2021zero}. For example, DALL-E~\cite{ramesh2021zero} trains a 12-billion-parameter autoregressive transformer for zero-shot text-to-image generation, and achievse the significant generalization. 

\subsection{Natural Language Processing}
With a long-term development, NLP has been developed into a broad sub-areas including tagging, named entity recognition, question answering and machine translations \textit{etc}. In the following, we briefly review three recently challenging sub-areas. For other pos tagging or text categorization tasks, we kindly refer to these literatures~\cite{akbik2018contextual,adhikari2019docbert}.

\begin{figure}[ht!] 
\begin{center}
\includegraphics[width=\linewidth]{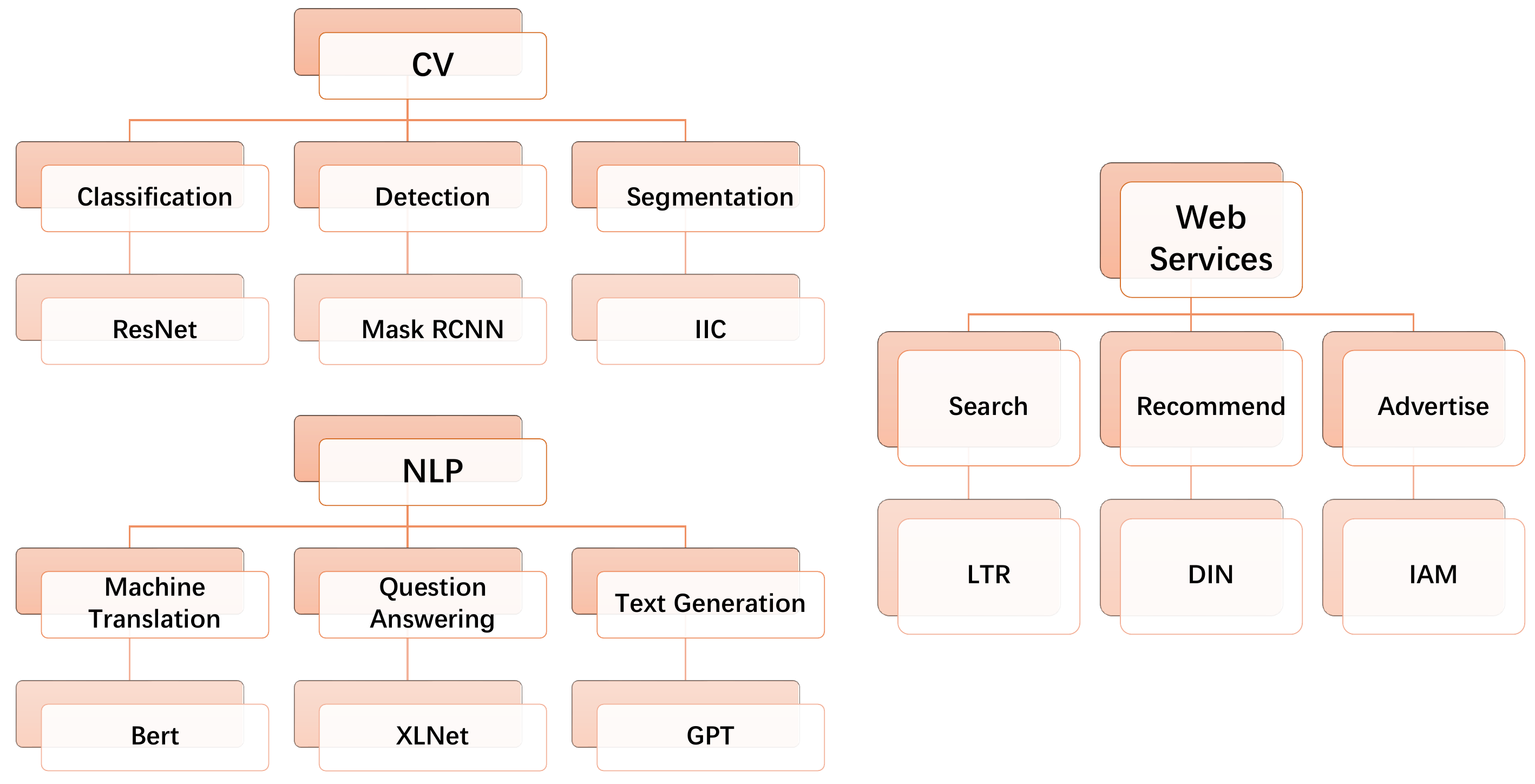}
\caption{\respond{We enumerate the representative methods at the bottom nodes of some subareas in CV, NLP and Web Services.} }\label{fig:cloud_exemplar}
\end{center} 
\vspace{-0.2in}
\end{figure}

Machine translation (MT)~\cite{hutchins1986machine} has achieved great success in movie caption translation, sport games and international political conference, which saves the human labor and domain knowledge in translation. Past works can be roughly divided into rule-based~\cite{simard2007rule}, statistical-based~\cite{koehn2009statistical} and neural machine translations~\cite{bahdanau2015neural}. The early attempt in neural MT is recurrent continuous translation model~\cite{kalchbrenner2013recurrent}, which leverages an autoregressive mechanism to automatically capture word ordering, syntax, and meaning of the source sentence explicitly. Sutskever \textit{et al.,}~\cite{sutskever2014sequence} improves it via an encoding-decoding approach of the bidirectional LSTM to learn the long-term dependencies. The following variants mainly target to design the more efficient attention modules. Recently, the BERT-based neural architecture achieves the new state-of-the-art on a range of MT benchmarks by enlarging the capacity with the large-scale corpora~\cite{zhu2019incorporating}.

Question Answering (QA)~\cite{hirschman2001natural} as a classical technique is commonly depolyed in some well-known engines like Microsoft Windows and Apple Siri. The industrial QA usually consists of multiple stages \textit{e.g.,} query recognition and expansion, answer selection and fine-grained ranking spanning from text, image and videos~\cite{toxtli2018understanding}. The mainstream research falls into modeling the QA pairs and the negative sampling, exploring the implicit matching between multiple objects and the answer~\cite{antol2015vqa}. Dialogue system can be considered as a more complex QA, which generates the answer in the multiple rounds~\cite{merdivan2019dialogue}. Compared to the single retrieval, the interactive sentence generation is more critical.

Text generation is an emerging direction along with the development of the large-scale PTMs such as GPT series in the recent years~\cite{radford2019language,floridi2020gpt}. Several generation tasks like poetry generation and story generation have shown impressive performances. In order to make the generation more human-like, the GAN-style~\cite{lin2017adversarial} and VAEs-style~\cite{serban2017hierarchical} have been respectively explored in PTMs to leverage their potential merits. Except totally structure-free generation, some knowledge-enhanced text generation that considers the text hint, constraint-aware and graph-based knowledge have been investigated~\cite{yu2020survey,Zhang_Tan_Yu_Zhao_Kuang_Liu_Zhou_Yang_Wu_2020}. Besides, some cross-domain text generation like Visual QA and reading comprehension are also explored to understand the multi-modalities~\cite{floridi2020gpt,m6}.

\subsection{Web Services}
Recommendation, Search and E-advertisement has been the successful business paradigms in web services~\cite{kosala2000web}. The corresponding Cloud AI models as the core components of these paradigms are widely explored in many enterprises including Google, Amazon and Microsoft~\cite{sadiku2014cloud}.

\subsubsection{Search}
Web search~\cite{broder2002taxonomy} retrieves the object of interest from a huge number of candidates according to the human query. The keyword search has been investigated almost along with the emergence of world wide web~\cite{liu2006effective}. More challenges raise when query is the image or the video. In the image search, the model should extract any possible instances from the image to match the candidates and thus many works focus on enhancing the feature extraction ability~\cite{gordo2016deep}. Considering the domain bias and noise, some approaches proposed to pretrain the model on the in-domain clean subset and then finetune on the open-domain dataset~\cite{liu2021image}. Due to the latency constraint for the multimedia data, the hash techniques are usually explored to accelerate the retrieval~\cite{liu2016deep}.

\subsubsection{Recommendation}
Recommender systems have been widely studied in the last decade and become an indispensable infrastructure of web services~\cite{resnick1997recommender}.  The related recommendation methods are progressively improved with the development of collaborative filtering, deep learning and sequential modeling. The early stage mainly focuses on the user-based collaborative filtering~\cite{zhao2010user}, the item-based collaborative filtering~\cite{sarwar2001item} and matrix factorization~\cite{koren2009matrix}. As deep learning achieved a great success, several variants of collaborative
filtering combined with DNNs were proposed~\cite{cheng2016wide,he2017neural,guo2017deepfm,cui2018variational,yao2017discovering,chen2020towards}. They leverage DNNs to activate high-level semantics for more accurate recommendation. Sequential modeling as another perspective to model the user interests has been successfully applied to recommender systems~\cite{shani2005mdp}. With the architecture evolving, several methods based on GRU~\cite{jannach2017recurrent}, Attention~\cite{kang2018self,zhou2018deep,tan2021sparse,zhang2021cause,Lu_Huang_Zhang_Han_Chen_Zhao_Wu_2021,pan2021click} have achieved remarkable performances. 

\subsubsection{Advertisement}
Advertisement we review here refers to the computational advertising developed for web services~\cite{huh2020advancing}. It is an intersection of multiple disciplines like Advertising, Marketing and Computer Science. Computational Advertising builds on the deep understanding of the web data and user preference, and at the same time taking the cost and the revenue into account. A range of works study the recommendation system for advertising~\cite{schafer2007collaborative}, guaranteed target display advertising~\cite{turner2012planning} in a dynamic way or real-time bidding~\cite{schwartz2017customer} by analyzing the contextual data. Except the cost-effective impression, most algorithms target to find the best strategic planing among the complex markets with the uncertain revenue, or explore the optimal advertising path by inferring the causal relations with the incentive bonus~\cite{chu2020inductive}. With the development of AI, models combined with advertising have drawn more attention. It constructs an automatic optimization for the complex marketing environment while still maintaining the advertisement efficiency compared to the traditional advertisement~\cite{yun2020challenges}.

 \section{Edge AI}
On the mobile platform, the breakthrough of AI has spawn a wealth of intelligent applications, such as virtual assistants\cite{hoy2018alexa} and personalized recommendation\cite{pimenidis2019mobile}. The traditional cloud-based paradigm requires the data uploading, which may violate the user privacy and heavily depend on the network conditions~\cite{shuja2021applying}. 
The way to alleviate the above problems is edge inference, where we can place models partially or fully on mobile devices and make predictions locally. However, edge inference is highly nontrivial as the computing, storage and energy resources of mobile devices are limited, and many research efforts have been devoted. In this section, we review some representative hardware and important techniques to ease models on the edges.

\subsection{AI Computing Hardware in Edge Computing}
The demand for edge AI grows rapidly due to the issues of bandwidth, privacy or compute-transmission balance. Many hardwares have been developed to meet the requirements of numerous edge AI applications \respond{including the light-weighted servers such as Raspberry Pi and Nvidia TX 2, and some function-oriented hardwares}. Due to the space limitation, we simply enumerate some representative types of the function-oriented hardwares in the following \respond{and more general servers for edge computing can refer to~\cite{wang2020convergence}}.

\subsubsection{VPU}
Vision processing unit (VPU) is an emerging class of microprocessor used in edge AI. It allows the efficient execution of edge vision workloads and achieves a balance between power supply efficiency and computing performance. One of the most popular examples is the Intel Neural Compute Stick, which is based on the Intel Movidius Myriad X VPU. This plug-in and play device can be easily attached to edges running Linux, Windows, Raspbian, including Raspberry Pi and Intel NUC \textit{etc}. In terms of machine learning framework it supports TensorFlow, Caffe, MXNet and PyTorch \textit{etc}. 

\subsubsection{Edge TPU}
Google has built Edge Tensor Processing Unit (Edge TPU)\footnote{https://cloud.google.com/edge-tpu} to accelerate the ML inference on edges. It is capable of running the classical compressed CNNs such as MobileNets, MobileNets SSD and Inception as well as TensorFlow Lite models, and has been applied into the real-world detection and segmentation. Currently, it is limited to open for public.

\subsubsection{Mobile GPU}
Enterprises like Nvidia and Qualcomm explore to integrate the GPU to accelerate the computation on edges. The representative is NVIDIA Jetson Nano\footnote{https://developer.nvidia.com/embedded/jetson-nano}, which includes an integrated 128-core Maxwell GPU, quad-core ARM A57 64-bit CPU, 4GB LPDDR4 memory, along with support for MIPI CSI-2 and PCIe Gen2 high-speed I/O. Apple also developed the processor A12 Bionic for iPhone and iPad, which is a 64-bit ARM-based system on a Apple chip. It includes a dedicated neural network hardware, which Apple calls the ``next generation neural engine". Similarly, HiSilicon released Kirin 990 5G, a 64-bit high-performance mobile ARM 5G SoC. More recently, Qualcomm announced the Snapdragon 888 on 2020, which combines the computing capabilities of the new Hexagon 780 and GPU, providing the impressive 26 TOPS computing power.

\begin{figure}[t!] 
\begin{center}
\includegraphics[width=\linewidth]{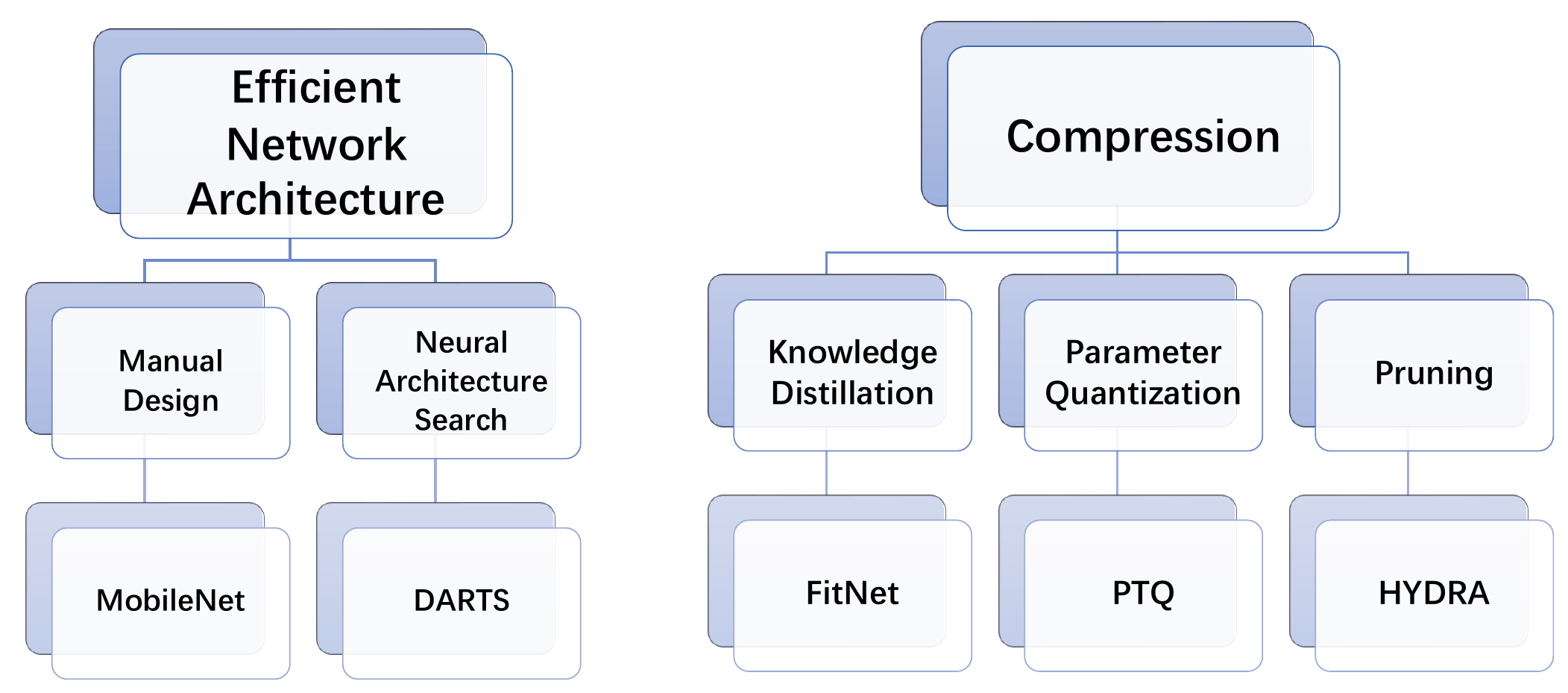}
\caption{\respond{We enumerate some representation methods in the area of Efficient Network Architecture and Compression.} }\label{fig:edge_exemplar}
\end{center} 
\vspace{-0.2in}
\end{figure}

\subsubsection{NPU}
Neural Processing Unit (NPU) is to find an efficient configuration to control a large number of resources~\cite{Park_Lee_Lee_Moon_Kwon_Ha_Kim_Park_Bang_Lim_2021,Lee_2021}. Typically, DNNs require a large amount of data for training and ask for large memory. To resolve the need for off-/on-chip memory bandwidth, NPUs usually rely on data reuse and unnecessary computations skipping. For example, ARM Ethos N-77\footnote{https://www.arm.com/products/silicon-ip-cpu/ethos/ethos-n77} delivers up to 4 TOPS of performance, scaling to 100s of TOPS in multicore deployments. It supports machine learning frameworks like TensorFlow, TensorFlow Lite, Caffe2, PyTorch,
MXNet, and ONNX. Tensorflow and Pytorch also have generic support for deploying models on Android NPU via Google NNAPI.

\subsection{Efficient Network Architecture}
\subsubsection{Manual Design}
There are a lot of works exploring the lightweight network architectures. One representative is SqueezeNet~\cite{hu2018squeeze}, which leverages the iteration of a squeeze layer and a expansion layer for the parameter compression. MobileNet~\cite{howard2017mobilenets} decomposes the conventional convolution into the composition of a depth-wise convolution and a point-wise convolution. The intuition behind MobileNet is that the low rank merit of the convolution kernel makes the decomposition approximately equivalent, and thus computation can be accelerated by a two-stage convolution. 
Similarly, MobileNet-v2~\cite{sandler2018mobilenetv2} shows that the high dimensional features can actually be expressed through compacting low-dimensional features. Then, they proposes a new layer ``inverted residual with linear bottleneck" to reduce the parameters. ShuffleNet~\cite{zhang2018shufflenet} shows the point-wise convolution in MobileNet is expensive when the input dimension is high. It leverages the group convolution together with the channel shuffle to reduce the computational cost and the parameter space.

\subsubsection{Neural Architecture Search}
Compared to the manual design,
neural architecture search (NAS) enables us to automatically explore the efficient network architectures. 
The classical NAS~\cite{zoph2016neural} utilizes RNN as the controller to generate a sub-network, and then perform the training and evaluation, and finally updates the parameters of the controller. 
There are two major challenges in NAS~\cite{brock2017smash,zoph2016neural,klein2017fast}. 
The first comes from the non-differential objective~\cite{elsken2018efficient,cai2018efficient,liu2018progressive}. Specifically, the performance of the sub-network is non-differential, which makes it infeasible to optimize the controller directly. 
Fortunately, the strategic gradient methods in RL can be a surrogate to update the controller parameters.
Another challenge relates to the computationally expensive pipeline where we have to train each sub-network updated by the controller from scratch~\cite{zoph2018learning,real2019aging,yu2019evaluating}.
Towards this end, efficient neural architecture search introduces the weight sharing technique and greatly reduces the search time~\cite{pham2018efficient}. Recently, there is a rapidly growing trend in this direction, which can be divided in the perspectives of search space~\cite{real2019aging}, search policy sampling network~\cite{ liu2018darts} and the performance-aware selection~\cite{yu2019evaluating}.

\subsection{Compression}
\subsubsection{Knowledge Distillation} 
KD~\cite{hinton2015distilling} is widely used to transfer the knowledge from complex models or model ensembles to the lightweight models. 
According to the distillation manner, KD can be divided into response-based KD~\cite{hinton2015distilling, zhang2018deep, cho2019efficacy, yang2019training}, feature-based KD~\cite{romero2014fitnets, zagoruyko2016paying} and relation-based KD~\cite{yim2017gift, tung2019similarity}.

\begin{itemize}[leftmargin=*]
    \item \textbf{Response-Based KD.} It takes the network output as the soft target to teach the student model.
    It is simple but effective for model compression, and has been widely used in different tasks and applications.
    \item \textbf{Feature-Based KD.} 
    DNNs are good at learning multiple levels of feature representation by abstraction. Therefore, the output of the intermediate layer, \textit{i.e.,}, the feature map, can be used as the knowledge to supervise the training of the student model. FitNet~\cite{romero2014fitnets} improved the training of the student model by matching the feature map between teachers and students directly. Subsequently, a range of other methods have been proposed to follow this paradigm~\cite{zagoruyko2016paying,kim2018paraphrasing,heo2019knowledge,passban2020alp,chen2021cross,wang2020exclusivity,48gan2020bert}. For example, Zagoruyko \textit{et al.,}~\cite{zagoruyko2016paying} derived an attention map from the original feature maps to express knowledge.
    
    \item \textbf{Relation-Based KD.} This lines of methods explore the relationship between different layers or data samples for distillation. Specifically, to explore the relationships between different feature maps, a flow of solution process (FSP) is proposed, which is defined by the gram matrix between two layers~\cite{yim2017gift}. 
    The FSP matrix summarizes the relations between pairs of feature maps.
    It is calculated using the inner products between features from two layers. To use the knowledge from multiple teachers, two graphs are formed by respectively using the logits and features of each teacher model as the nodes~\cite{zhang2018better,lee2019graph}. 
\end{itemize}

\begin{figure*}[ht!] 
\centering
\begin{subfigure}[b]{.33\textwidth}
\centering
\includegraphics[width=\textwidth]{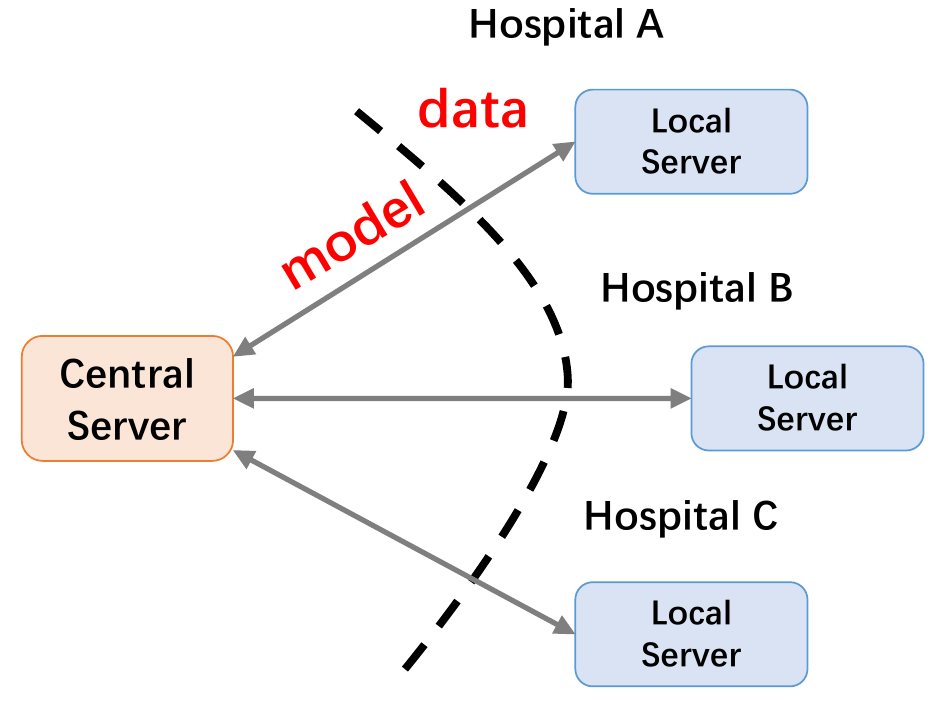}
\caption{Privacy-Primary Collaboration}
\label{fig:prototypes-privacy}
\end{subfigure}
\begin{subfigure}[b]{.33\textwidth}
\centering
\includegraphics[width=\textwidth]{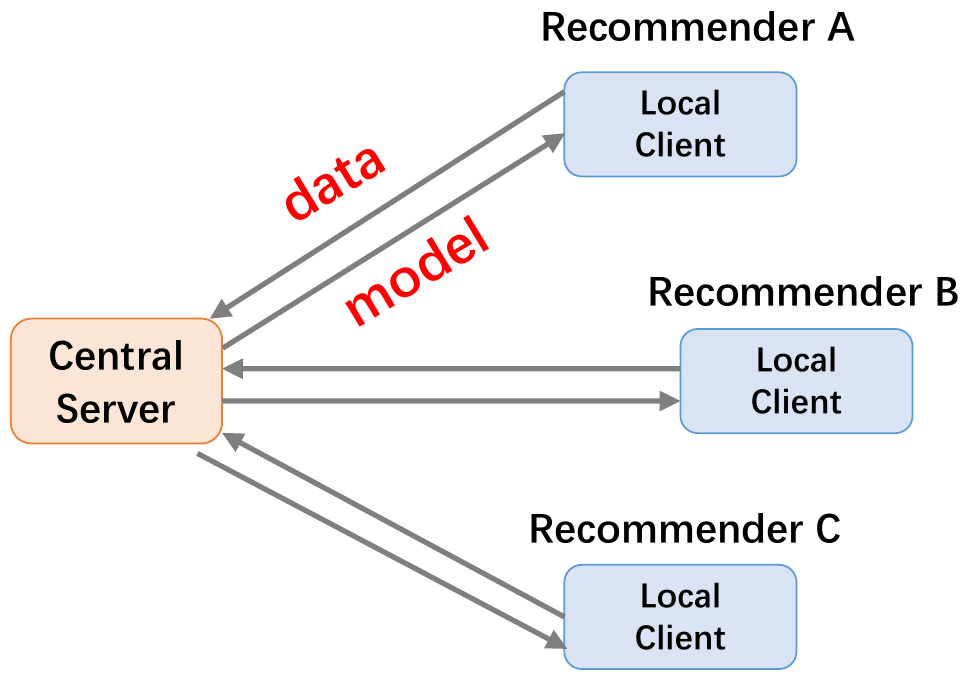}
\caption{Efficiency-Primary Collaboration (I)}
\label{fig:prototypes-efficiency}
\end{subfigure}
\begin{subfigure}[b]{.33\textwidth}
\centering
\includegraphics[width=\textwidth]{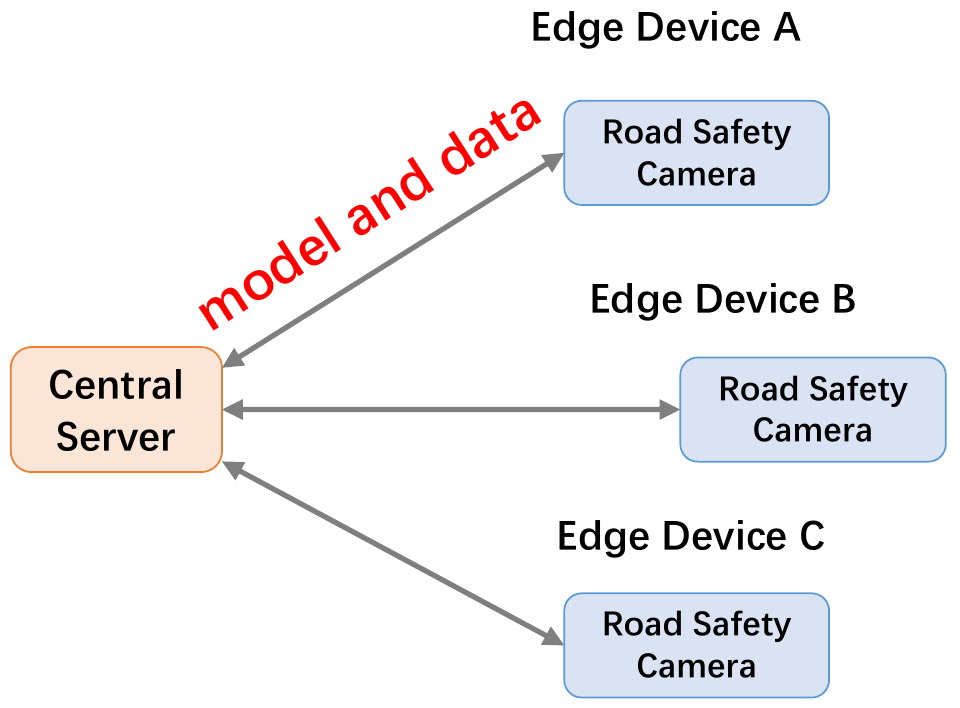}
\caption{Efficiency-Primary Collaboration (II)}
\label{fig:prototypes-efficiency2}
\end{subfigure}
\caption{\respond{For each prototype of Edge-Cloud Collaborative AI mentioned in the paragraph, we present an instance. Specifically, in privacy-primary collaboration (a) \textit{e.g.,} federated learning, the patient data in the hospital is very private, and thus during training and serving, only the model parameters are allowed to communicate. In efficiency-primary collaboration (I) \textit{e.g.,} EdgeRec, the fine-grained user behavior data is collected to the central server, which trains on-device recommendation (personalized) models and deploys to clients. Finally, in efficiency-primary collaboration (II), both central server and local clients will respectively maintain independent models which interact with each other over data and model features.}}
\label{fig:prototypes}
\vspace{-0.2in}
\end{figure*}

\subsubsection{Parameter Quantization} 
Quantization has shown a great success in both training and inference phases of DNN models~\cite{banner2018scalable,han2016deep}. Specifically, the breakthrough of the half-precision and mixed-precision
training~\cite{micikevicius2017mixed,gupta2015deep,ginsburg2017tensor,courbariaux2014training,banner2018scalable,chmiel2020neural,faghri2020adaptive,li2019additive} has enabled an order of magnitude higher throughput in AI accelerators. However, it has proven very difficult to go below half-precision without significant tuning, and thus most of the recent research has focused on Parameter Quantization (PQ)~\cite{han2016deep}. Currently, the PQ methods can be divided into Quantization-Aware Training and Post-Training Quantization.
\begin{itemize}[leftmargin=*]
    \item \textbf{Quantization-Aware Training (QAT):} In QAT, the forward pass and the backward pass are performed on the quantized model in floating point, and the model parameters are quantized after each gradient update. Performing the backward pass with floating point is important, since accumulating the gradients in quantized precision can result in the zero gradient or gradients that have high error, especially in low-precision~\cite{courbariaux2015binaryconnect,gysel2016hardware,gysel2018ristretto,huang2018data,lin2015neural,rastegari2016xnor}.
    A popular approach to address this is to approximate the gradient by the Straight Through Estimator (STE)~\cite{bengio2013estimating}.
    However, the computational cost of QAT is usually very high, since the retraining of the model may take hundreds of epochs, especially for the low bit quantization.
    \item \textbf{Post-Training Quantization (PTQ):} An alternative is Post-Training Quantization, which performs quantization and adjustment of the weights without fine-tuning~\cite{banner2018post,cai2020zeroq,choukroun2019low,fang2020post,garg2021confounding,he2018learning,hubara2020improving,lee2018quantization,zhao2019improving}. The overhead of PTQ is often negligible, making PTQ appliable in situations where data is limited or unlabeled. However, this often comes at the cost of the lower accuracy compared to QAT, especially for the low-precision quantization.
    In PTQ, all the quantization parameters are determined without any re-training of the DNN model.
\end{itemize}

\subsubsection{Pruning}
Pruning is another way to reduce the parameter space by removing some computation paths in the model. Previous methods in this direction can be categorized into two main categories, one-time pruning and runtime pruning~\cite{vysogorets2021connectivity}. Specifically, for the former, there are three lines. One line of methods focus on pruning after model training~\cite{lecun1990optimal,hassibi1993optimal,han2015learning,dong2017learning}. It aims to design the certain criteria like value magnitudes~\cite{han2015learning} and second-derivative information~\cite{lecun1990optimal} to remove the least salient parameters. Another line of methods~\cite{narang2017exploring,zhu2017prune,mocanu2018scalable,dettmers2019sparse,evci2020rigging} focus on jointly learning the sparse structures with model parameters by considering different downstream hardware features, \textit{e.g.,} OFA~\cite{cai2020once}. These works explore the efficient strategies to avoid the expensive prune-retrain cycles for diverse constraints. The third line aims to prune networks in initialization, which saves the training resources~\cite{lee2018snip,wang2019picking,tanaka2020pruning}. The Lottery Ticket Hypothesis~\cite{frankle2018lottery} demonstrates that the dense network contain sub-networks, that can reach a comparable accuracy with the full network. 

Runtime pruning focus on the dynamic path selection to enable the real-time inference with the limited computation budgets. For example, Huang \textit{et al.,}~\cite{huang2018multi} implements the early-exit mechanism by injecting a cascade of intermediate classifiers throughout the deep CNNs. The layers are evaluated one-by-one with the forward process and the early-stop will execute once the CPU time is depleted, namely the latter layers are pruned. Yu \textit{et al.,}~\cite{yu2018slimmable} introduces a slimmable neural network, which adjusts its width on the fly based on the resource constraints. In contrast to the depth pruning, reducing width helps reduce the memory footprint in inference. Besides, some others re-consider the input-dependent assumption, since treating all inputs equally may cause unnecessary resource consumption~\cite{teerapittayanon2016branchynet,huang2018multi,bolukbasi2017adaptive,wang2018skipnet,wu2018blockdrop,lin2017runtime}. For inputs that are easy to distinguish, a simpler model might be sufficient. The pioneering works use handcrafted control decisions~\cite{teerapittayanon2016branchynet,huang2018multi}. For instance, Huang \textit{et al.,}~\cite{huang2018multi} terminates the inference once the intermediate classifiers output the confidence score exceeding a pre-determined threshold. 
The subsequent improvements propose to learn a network selection system, which adaptively prunes the full network for each input example. For example, Bolukbasi \textit{et al.,}~\cite{bolukbasi2017adaptive} propose an adaptive early-exit strategy by introducing extra classifiers to determine whether the current example should proceed to the next layer. Some others~\cite{wang2018skipnet,wu2018blockdrop,lin2017runtime} utilize RL to learn the dynamic pruning decisions, which has a higher structure variability compared to the early-exit mechanism.

\section{Edge-Cloud Collaborative AI}
Edge-cloud collaborative modeling has received significant interest across separate research communities \respond{due to the recent privacy concerns, the latency requirements or the personalization~\cite{zhou2019edge}. For example, for the patient data, it is private and sensitive to open for the training and serving of AI models. This motivates the development of federated learning, which distributes the training to the local server for the purpose of privacy and aggregates the final model in the cloud server. From this perspective, federated learning is a classical edge-cloud ``collaboration" that makes AI more practical. Actually, there are some similar exemplars about the different forms of collaboration and to sum all, } we can categorize them into two sets: privacy-primary collaboration and efficiency-primary collaboration (\textit{cf.} Figure \ref{fig:prototypes}).

\subsection{Privacy-Primary Collaboration: Federated Learning}
As aforementioned above, the collaboration of FL is privacy-primary in the sense that raw data of each client is stored locally and will not be exchanged or transferred~\cite{kairouz2019advances}.
Typically, FL can be roughly divided into two lines of works, \textit{i.e.}, \textit{cross-device} and \textit{cross-silo}, respectively. In cross-device FL, there are substantial devices like phones, laptops, or IoT. In cross-silo FL, differently, there are organizations where data silos naturally exist.
Another categorization given by \cite{yang2019federated} divides FL into horizontal FL, vertical FL and federated transfer learning, according to how data of different participants overlaps. 
In modern FL frameworks, there are several challenges that hinder effective edge-cloud collaboration. In this paper, we focus on three major challenges, \textit{i.e.}, data heterogeneity, communication efficiency and attacks on FL.

\subsubsection{Statistical Heterogeneity}
The real-world data samples are not usually independent and identically distributed over numerous devices~\cite{li2020federated}, resulting in heterogeneous edge models and gradients. 
Typically, there are five sources of data heterogeneity: 
1) \emph{Feature skew}: The marginal distribution of features may differ between edges. 
2) \emph{Label skew}: The marginal distribution of labels may differ between edges.
3) \emph{Same label, different features}: The conditional distribution of features given labels
    may differ between edges.
4)  \emph{Same features, different labels}: The conditional distribution of labels given features
    may differ between edges.
    For example, the symbol $\checkmark$ represents \texttt{correct} in many countries and \texttt{incorrect} in some others (\textit{e.g.}, Japan);
and 5) \emph{Quantity skew}: Clients can store drastically varying volumes of data. 
Unfortunately, real-world FL datasets present a mix of these phenomenons, and it is an urgent need to drive measures to alleviate statistical heterogeneity for edge-cloud collaboration.
In this nascent research area, FL personalization addresses this issue by introducing a two-stage collaboration framework.
The first stage is to collaboratively build a global model, followed by a customization stage for each customer with the client's private data \cite{sim2019investigation}. 
Recently, several methods have been proposed to achieve personalization, enhanced by transfer learning, meta training, and KD techniques. A learning-theoretic framework with generalization guarantees is presented by \cite{mansour2020three}. Wang et al. \cite{wang2019federated} construct different personalization strategies, including the model graph, and the training hyperparameters for different edges.
Jiang et al. \cite{jiang2019improving} build connections between FL and meta-learning, and interpret FedAvg as a popular algorithm, Reptile \cite{nichol2018first}.
Shen et al. \cite{shen2020federated} presents an FL method based on KD and transfer learning that allows clients to train their own model independently with the local private data. 
Instead of training one global model, Masour et al. \cite{mansour2020three} consider device clustering and learn one model per cluster.

\subsubsection{\respond{Communication Efficiency}}
\respond{Communication cost is an important factor when deploying FL to the real-world scenarios given the limited communication budget. Many algorithms~\cite{hamer2020fedboost,hou2021reducing,liu2020client} have explored to improve the communication efficiency by accelerating the convergence rate of the training model or introducing a hierarchical counterpart of federated learning. Specially, the latter perspective is scalably in many practical cases by introducing the intermediate servers to perform the partial aggregation. The subsequent improvement regarding the dynamic resource allocation and the task management to save the transmission cost in this hierarchy are also explored~\cite{lim2021decentralized,lim2021dynamic}. In the future, with the development of the unmanned aerial vehicles, it is possible to achieve the near-instant super-connectivity for federated learning in a more communication-efficient manner~\cite{lim2021uav,lim2021towards}.
}

\begin{table*}[!ht]
    \centering
    \caption{Device-Cloud Collaborative Modeling}
    \begin{tabular}{l|cc|cccc}
        \toprule
         & \multicolumn{2}{l|}{\textbf{Collaboration  Manner}} & \multicolumn{4}{l}{\textbf{Collaboration Concerns}} \\ 
         \cmidrule{2-7}
         & co-training & co-inference & privacy & storage efficiency& communication efficiency& personalization\\ 
         \midrule
        Semantic QA cache~\cite{yoon2016device} &  & \checkmark &  &  & \checkmark & \checkmark  \\ 
        EdgeRec System~\cite{edgerec} &  & \checkmark &  &  & \checkmark & \checkmark  \\ 
        Auto-Split~\cite{banitalebi2021auto} &  & \checkmark &  & \checkmark & \checkmark &   \\ 
        CoEdge~\cite{hu2020coedge} &  & \checkmark &  & \checkmark & \checkmark &   \\ 
        Colla~\cite{lu2019collaborative} & \checkmark &  &  & \checkmark &  & \checkmark  \\ 
        DCCL~\cite{dccl} & \checkmark &  &  & \checkmark &  & \checkmark  \\ 
        MC$^2$-SF~\cite{chen2021mc} & \checkmark & \checkmark &  &  &  & \checkmark  \\ 
        FedAvg~\cite{mcmahan2017communication} & \checkmark &  & \checkmark &  & \checkmark &   \\ 
        FML~\cite{shen2020federated} & \checkmark &  & \checkmark & \checkmark & \checkmark & \checkmark  \\ 
        Personalized FedAvg~\cite{jiang2019improving} & \checkmark &  & \checkmark &  & \checkmark &\checkmark  \\
        HyperCluster~\cite{mansour2020three} & \checkmark &  & \checkmark &  & \checkmark & \checkmark  \\
        Federated Evaluation~\cite{wang2019federated} & \checkmark & \checkmark & \checkmark &  &  & \checkmark \\
        \bottomrule
    \end{tabular}
    \label{Collaboration}
\end{table*}

\subsubsection{Attacks on FL}
Current FL protocol designs are usually vulnerable to attackers inside and outside of the system, yielding the privacy and the robustness at risk.
There are two serious threats to FL privacy and robustness \cite{lyu2020privacy}: 1) poisoning attacks against robustness; and 2) inference attacks against privacy.

The impact of \textit{poisoning attacks} on the FL model is determined by the extent to which the backdoor players participate in the assaults, as well as the amount of training data poisoned. Model poisoning attacks seek to prevent global model learning \cite{lamport2019byzantine} or hide a backdoor trigger into the global model \cite{wang2020attack}. These attacks contaminate the changes of local models before these changes are uploaded to the server. The Byzantine attack \cite{lamport2019byzantine} is a form of untargeted model poisoning attack that uploads arbitrary and harmful model changes to the server that fools the global model. 

Though FL protects data privacy by shielding off local data from being directly accessed, FL still suffers from privacy risks due to \textit{inference attacks}. For example, Deep Leakage from Gradient~\cite{zhu2020deep} presents an optimization approach that can retrieve the raw images and texts critical for model improvement.
Existing studies in privacy-preserving FL are often built on traditional privacy-preserving approaches, including: (1) \emph{homomorphic encryption}\cite{paillier1999public}; (2) \emph{Secure Multiparty Computation}, \cite{demmler2015aby}; and (3) \emph{differential privacy}~\cite{dwork2014algorithmic}. 
To protect against an honest-but-curious opponent, Hardy et al. \cite{hardy2017private} used FL on data encrypted by the homomorphic technique. Yao et al.~\cite{yao1982protocols} introduces a collaborative computation on their inputs without disclosing them to one another. 
Given the resource limits of mobile devices, it is expected that privacy-protection solutions must be computationally inexpensive, communication-efficient and resistant to failure. 
Truex et al.~\cite{truex2020ldp} present a FL system under the protection of the local differential privacy framework, while minimizing the overwhelming impact of noise.

\subsection{Efficiency-Primary Collaboration}
Although FL is a popular framework that considers the data privacy and governance issue, some real-world applications may not be sensitive to the privacy but care more about the factors like communication budget or personalization. In the recent years, there are some emerging edge-cloud collaborations in this direction and we term this paradigm as the \emph{efficiency-primary collaboration} (\textit{cf.} Figure \ref{fig:prototypes-efficiency}-\ref{fig:prototypes-efficiency2}).

\subsubsection{\respond{Split-deployment}}
The direct collaboration is the \respond{partially inference offloading from cloud to edge or from edge to cloud~\cite{dey2019offloaded,xu2020energy,pacheco2021calibration,pacheco2021early}}, which separates one model into two components, one on the cloud side and the other placed on the edge side~\cite{zhou2021machine,shuja2017analysis}. We term it as the \emph{split-deployment}. One successful practice is from the Taobao EdgeRec System~\cite{edgerec}, where the memory-consuming embedding matrices encoding the attributes are deployed on the cloud side and the lightweight component executes the remaining inference on the edge side. Amin \textit{et al.,}~\cite{banitalebi2021auto} introduced an Auto-Split solution to automatically split DNNs models into two parts respectively for the edge and for the cloud. \respond{The coded edge computing based on the auction mechanism~\cite{auction1,auction2} is also a promising method to achieve the efficient resource allocation considering the edge capacities.} To optimize the latency, \cite{li2018jalad}\cite{ko2018edge} leverage lossy data compression techniques to reduce the transmission data size. \respond{Some scenarios involve in the unmanned aerial vehicles as the assisted edge-cloud computing, which requires a more complete joint learning paradigm for the task allocation and management~\cite{uav}.} 

\subsubsection{Edge-Centralized Personalization}
Another type of edge-cloud collaboration is to leverage the decentralized advantage of the edge side in personalization by setting up an auxiliary cloud model. For example, Lu~\textit{et.al.,}~\cite{lu2019collaborative} proposes a collaborative learning method, COLLA, for the user location prediction, which builds a personalized model for each device, and allows the cloud and edges learned collectively. In COLLA, the cloud model acts as a global aggregator distilling knowledge from multiple edge models. Ding~\textit{et.al.,}~\cite{ding2020cloud} introduces a collaborative framework, where a CloudCNN provides the soft supervision to each local EdgeCNN for the edge training and simultaneously, the EdgeCNN performs the real-time inference interacting with the vision input. Yao~\textit{et.al.,}~\cite{dccl} proposes a edge-cloud collaboration framework for recommendation via the backbone-patch decomposition. It greatly reduces the computational burden of edges and guarantees the personalization by introducing a MetaPatch mechanism and re-calibrates the backbone to avoid the local optimum via MoModistill. Extensive experiments demonstrate its superiority on the recommendation benchmark datasets.

\subsubsection{Bidirectional Collaboration}
One more intensive type of edge-cloud collaboration could be bidirectional, where we consider the independent modeling on each side and they maintain the interactive feedback to each other during both the training and serving. One recent exploration is a Slow-Fast Learning mechanism for Edge-Cloud collaborative recommendation developed on Alibaba Luoxi Platform~\cite{chen2021mc}. In this method, the slow component (the cloud model) helps the fast component (the edge model) make predictions by delivering the auxiliary latent representations; and conversely, the fast component transfers the feedbacks from the real-time exposed items to the slow component, which helps better capture user interests. The intuition resembles the role of System I and System II in the human recognition~\cite{kahneman2011thinking}, where System II makes the slow changes but conducts the comprehensive reasoning along with the circumstances, and System I perceives fast to make the accurate recognition~\cite{madan2021fast}. The interaction between System I and System II allows the prior/privileged information exchanged in time to collaboratively meet the 
requirements of the environment. With the hardware advancement of mobile phones, IoT devices and edge servers, it will be meaningful to pay attention to such a bidirectional collaborative mechanism in different levels~\cite{zhou2019edge}.

\subsection{Rethinking Collaboration in Classical Paradigms}

Edge-cloud collaborative learning can be formulated as a two-stage optimization problem, where we train a model on one side (cloud or edge) and then further optimize it on the other side (edge or cloud). Considering this, we rethink this problem from the perspectives of transfer learning, meta-learning, and causal inference.

\subsubsection{Transfer Learning.} 
In edge-cloud collaborative learning, the data distribution naturally differs from edge to edge and from edge to cloud. Towards this end, heterogeneous transfer learning \cite{Zhang_Qi_Yang_Prisacariu_Wah_Torr_2020,Yeh_Huang_Wang_2014,Wang_Wu_Jia_2017,Ren_Feng_Dai_Yan_2021,Wu_Zhu_Yan_Wu_Zhang_Ng_2021,Tsai_Yeh_Wang_2016,Samat_Persello_Gamba_Liu_Abuduwaili_Li_2017a,Li_Wang_Zhang_Li_Keutzer_Darrell_Zhao_2021} would greatly improve the bidirectional model adaptation between the edges and the cloud.
There are roughly two lines of heterogeneous transfer learning works addressing the difference in feature space, \textit{i.e.}, symmetric transformation, and asymmetric transformation. Symmetric transformation \cite{Samat_Persello_Gamba_Liu_Abuduwaili_Li_2017a,Wang_Ma_Cheng_Zou_Rodrigues_2018,Yeh_Huang_Wang_2014} aims to learn domain-invariant representations across different domains. \cite{Liu_Zhang_Lu_Lu_2017} addressed unsupervised transfer learning, which indicates a mostly labeled source domain with no target domain labels. 
\cite{Tsai_Yeh_Wang_2016} proposed Cross-Domain Landmark Selection as a semi-supervised solution. 
As a counterpart, by asymmetric transformation mapping \cite{Zhou_Tsang_Pan_Tan_2014,Kulis_Saenko_Darrell_2011,Feuz_Cook_2015,Xiao_Guo_2015a}, the feature space of the source domain is aligned with that of the target domain. 
A semi-supervised method of adapting heterogeneous domains was proposed, called Semi-Supervised Kernel Matching Domain Adaptation~\cite{Xiao_Guo_2015a}. 
\cite{Wu_Wu_Ng_2022} learns an enhanced feature space by jointly minimizing the information loss and maximizing the domain distribution alignment.

\subsubsection{Meta-learning.} Meta-learning is another successful knowledge transfer framework. Different from transfer learning where models learn from solving the tasks in the source domain, meta-learning expects that the model learns how to quickly solve new tasks. 
Based on meta-learning, on-cloud training (meta-training) could yield models that can learn quickly in heterogeneous edge environments (meta-learning). Recently, \cite{Rosenfeld_Rajendran_Simeone_2021} proposed to employ the spiking neural networks and meta-learning with streaming data, permitting fast edge adaptation.
Based on MAML \cite{finn2017model}, MELO \cite{Huang_Zhang_Yang_Qian_Wu_2021} is another work that learns to quickly adapt to new mobile edge computing tasks. Another major challenge in edge-cloud collaborative learning is the limited computation capacity of edges. By consolidating both meta-learning and model compression, existing researches learn light-weight models that can quickly adapt to the edge environments \cite{Zhou_Xu_McAuley_2021,Ye_Zhang_Wang_2021,Pan_Wang_Qiu_Zhang_Li_Huang_2021,Zhang_Wang_Gai_2020}. \cite{Ye_Zhang_Wang_2021} proposes an end-to-end framework to seek layer-wise compression with meta-learning. 
\cite{Pan_Wang_Qiu_Zhang_Li_Huang_2021} learns a meta-teacher that is generalizable across domains and guides the student model to solve domain-specific tasks.

\subsubsection{Causal Inference.} There is a substantial and rapidly-growing research literature studying causality~\cite{kuang2020causal} for bias reduction and achieving fairness.
Causal theory is essential to edge-cloud collaborative learning for the following two reasons: 1) on-cloud training is supposed to yield a generalizable model, which is free from confounding effects and model bias, \textit{w.r.t} the heterogeneous edge data distributions; 2) edge training should avoid over-fitting plagued by spurious correlations between the input and the outcome. \cite{Rotman_Feder_Reichart_2021} builds connections among model compression, causal inference, and out-of-distribution generalization. With the causal effects as the basis, they propose to make decisions on model components pruning for better generalization. More works dive into the intersection of causality and out-of-domain generalization \cite{Teshima_Sato_Sugiyama_2020,Yang_Shen_Chen_Li_2020,Chen_Bhlmann_2020,Yue_Sun_Hua_Zhang_2021,kuang2018stable,yuan2021learning,Yang_Yu_Cao_Liu_Wang_Li_2020,zhang2020devlbert}. \cite{Yue_Sun_Hua_Zhang_2021} proposes to reserve semantics that is discriminative in the target domain by embracing disentangled causal mechanisms and deconfounding. \cite{Yang_Yu_Cao_Liu_Wang_Li_2020} assumes the relationship between causal features and the class is robust across domains, and adopts the Markov Blanket \cite{Yu_Guo_Liu_Li_Wang_Ling_Wu_2020} for the causal feature selection. 
\cite{kuang2018stable} proposes a causal regularizer to recover the causation between predictors and outcome variables for stale prediction across unknown distributions.
\cite{yuan2021learning} 
designs an instrumental variable based methods for achieving invariant relationship between predictors and outcome variable for domain generalization.

\section{Advanced Topics}
\subsection{Edge Pretraining Models}
PTM, also known as the \emph{foundation model}s~\cite{FoundationModel}, such as BERT~\cite{devlin2019bert} and GPT-3~\cite{brown2020language}, have become an indispensable component of modern AI, due to their versatility and transferability especially in few-shot or even zero-shot settings.
The scale of the foundation models have been growing tremendously recently~\cite{megatronlm,gshard,switchtrans,m6,m6-t}, as researchers observe that larger models trained on larger corpora generally leads to superior performance on downstream tasks~\cite{devlin2019bert,brown2020language}.
The large scale, which can range from millions of to trillions of parameters, however, raises a critical question as to whether edge intelligence can enjoy the merits of the increasingly powerful foundation models.
On the other hand, the unique challenges faced by edge agents, such as
the heterogeneity of the deployment environments and the need of small sample adaptation, constitute an ideal testbed for demonstrating the
versatility and transferability of the foundation models.
We herein discuss three crucial directions that require exploration before we can unleash the power of the foundation models on edges.

\subsubsection{Model Compression}
Compressing a large foundation model into a small one is necessary due to the storage and/or network bandwidth constraint of many edge mobiles such as modern mobile phones.
General techniques discussed in previous sections, such as model quantization and weight pruning, can be readily adopted, while there are also techniques specifically tailored for the Transformer architecture commonly adopted by the foundation models.
For instance, parameter sharing across layers has been proven effective by ALBERT~\cite{Lan2020ALBERT:}.
KD~\cite{hinton2015distilling} remains the most popular solutions.
DistilBERT~\cite{DistilBERT} uses a subset of the layers of the teacher model to form the student for distillation.
TinyBERT~\cite{jiao-etal-2020-tinybert} improves upon the vanilla logit-based distillation method~\cite{hinton2015distilling} by adding extra losses that align the immediate states between the teacher and the student.
MobileBERT~\cite{sun-etal-2020-mobilebert} introduces bottleneck layers to reduce the dense layers' parameters while keeping the projected hidden states' dimension unchanged, for convenient feature map transfer and attention transfer.
MiniLM~\cite{MiniLM} demonstrates that it is effective to distill the dot products between the attention queries, keys, and values, which do not impose constraints on the hidden size or the number of layers of the student.
So far the distilled tiny models usually comes with around 10M parameters, and it is unclear whether this size can be further reduced without much performance loss.

\subsubsection{Inference Acceleration}
Some techniques for reducing the model size, for example reducing the hidden sizes or the number of layers, can also bring the faster inference. Yet, as a foundation model typically has tens of layers, an interesting question arises: whether it is necessary to use all the layers during inference, since there may be simple samples or easy downstream tasks that do not necessitate using the full model.
PABEE~\cite{zhou2020bert} demonstrates that \emph{early exiting}, i.e.\ dynamically stopping inference once the intermediate predictions of the internal classifier layers remain unchanged for a number of steps, is indeed effective with BERT. However, it remains unclear whether early exiting is equally applicable to generative models such as GPT-3 and text generation tasks.

\subsubsection{Few-sample and Few-parameter Adaptation}
The traditional method is to fine-tune a PTM on each target downstream task's samples.
However, fine-tuning typically involves updating almost all parameters of a foundation model and is not effective enough in terms of sample efficiency~\cite{wei2021finetuned}.
It can lead waste of storage and network bandwidth since each fine-tuned model requires independent resources.
Moreover, an edge is typically a few-sample learning environment, for example, only consisting of data associated with one single mobile phone.
We thus need to seek a more efficient approach in place of fine-tuning for few-sample and few-parameter adaptation.
The recent emergence of prompt-based learning~\cite{gao2021making} and instruction tuning~\cite{wei2021finetuned} represent promising directions.
These recent methods mine natural language templates for the downstream tasks and use the templates to form input sentences.
With a powerful generative language model, such templates can guide the model to output the correct predictions for the downstream tasks, where the model prediction is in the form of natural language as well.
These recent paradigms require no parameter updating and can even achieve zero-shot learning in some cases~\cite{wei2021finetuned}.
However, so far prompt-based learning and instruction tuning focus mostly on large-scale models, not the type of tiny models for edges.

\subsection{Edge Graph Neural Networks}
In recent years, Graph Neural Networks have achieved the state-of-the-art on graph-structured data in various industrial domains~\cite{wu2020comprehensive}, including CV~\cite{yang2018graph,landrieu2018large}, NLP~\cite{marcheggiani2018exploiting,beck2018graph}, traffic~\cite{yao2018deep,li2017diffusion}, recommender systems~\cite{wu2019session} and chemistry~\cite{duvenaud2015convolutional}. 
It can learn high-level embedding from node features and adjacent relationship, thus effectively deal with graph-based tasks. 
The rapid growth of node feature and their adjacent information drive the success of GNN, but also pose challenges in integrating GNN into the edge-cloud collaboration framework, like data isolation, memory consuming, limited samples and generalization\textit{etc}.
Recently, some efforts have emerged to address the above problems from aspects of FL, quantization and meta learning. In this part, we briefly review these works.

\subsubsection{Federated GNN}
To collaborate the graph data distributed on different edges to train a high-quality graph model, recent researchers have made some progress in FL on GNN~\cite{jiang2020federated,zhou2020privacy,zheng2021asfgnn,wang2020graphfl,wu2021fedgnn,he2021fedgraphnn,wang2021fl,caldarola2021cluster,chen2021fedgl,he2021spreadgnn,meng2021cross,ni2021vertical,xie2021federated,zhang2021subgraph}. The key idea of 
FL is to leave the data on the edges and train a shared global model by uploading and aggregating the local updates, \textit{i.e.,} model parameters, to a central server.

For example, Feddy~\cite{jiang2020federated} proposes a distributed and secure framework to learn the object representations from multi-device graph sequences in surveillance systems.
ASFGNN~\cite{zheng2021asfgnn} further proposes a separated-federated GNN model, which decouples the training of GNN into two parts: the message passing part that is done by clients separately, and the loss computing part that is learnt by clients federally.
FedGNN~\cite{wu2021fedgnn} applies the federated GNN to the task of privacy-preserving recommendation. It can collectively train GNN models from decentralized user data and meanwhile exploit high-order user-item interaction information with privacy well protected.
FedSage~\cite{zhang2021subgraph} studies a more challenging yet realistic case where cross-subgraph edges are totally missing, and designs a missing neighbor generator with the corresponding local and federated training processes.
FL-AGCNs~\cite{wang2021fl} considers the NAS of GNN for the FL scenarios with distributed and private data.

\begin{figure}[t!]
    \centering
    \includegraphics[width=0.49\textwidth]{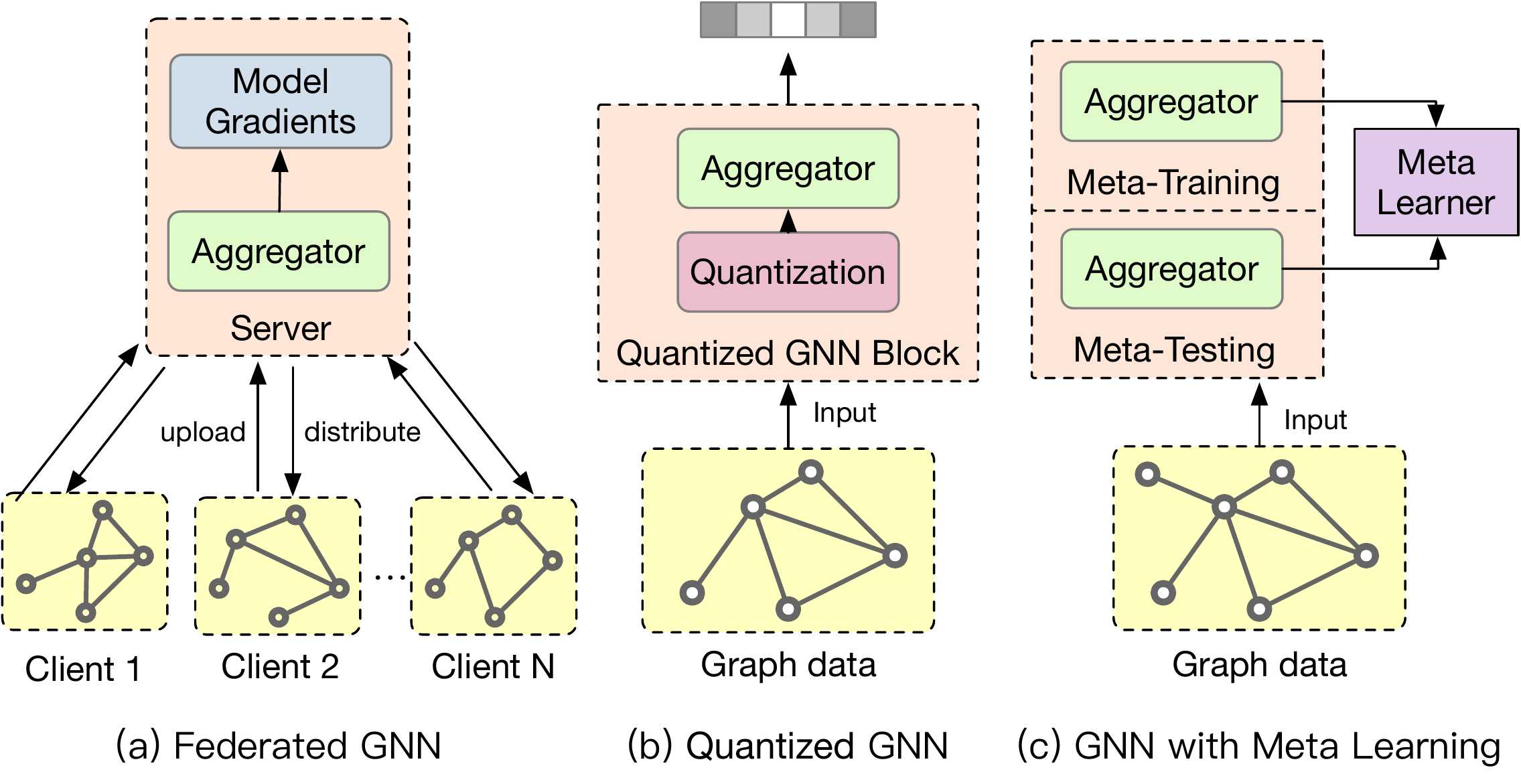}
    \caption{The ongoing evolution of Edge GNNs. (a) Federated GNN, (b) Quantized GNN, (c) GNN with Meta Learning.}
    \label{fig:my_label}
\end{figure}

To alleviate the heterogeneity in graph data, some works, e.g., FedCG~\cite{caldarola2021cluster} and GCFL~\cite{xie2021federated},
leverage clustering to reduce statistical heterogeneity by identifying homogeneous, while FedGL~\cite{chen2021fedgl} exploits the global self-supervision information.
SpreadGNN~\cite{he2021spreadgnn} extends federated multi-task learning to realistic serverless settings for GNNs, and utilizes a novel optimization algorithm with a convergence guarantee to solve decentralized multi-task learning problems.
In addition to data, graphs can also exist as relationships among clients. For example,
CNFGNN~\cite{meng2021cross} leverages the underlying graph structure of decentralized data by proposing a cross-node federated GNN. It bridges the gap between modeling complex spatio-temporal data and decentralized data processing by enabling GNN in FL.

\subsubsection{Quantized GNN}
To systematically reduce the GNN memory consumption, GNN-tailored quantization that converts a full-precision GNN to a quantized or binarized GNN can emerge as a solution for resource-constrained edge devices~\cite{feng2020sgquant,tailor2020degree,wang2021binarized,bahri2021binary,wang2021bi}. 
SGQuant~\cite{feng2020sgquant} proposes a multi-granularity quantization featured with component-wise, topology-aware, and layer-wise quantization to intelligently compress the GNN features while minimizing the accuracy drop.
Degree-Quant~\cite{tailor2020degree} performs quantization-aware training on graphs, which results in INT8 models often performing as well as their FP32 counterparts. 
BGN~\cite{wang2021binarized} learns binarized parameters and enables GNNs to learn discrete embedding. 
Bi-GCN~\cite{wang2021bi} binarizes both the network parameters and the node attributes and can significantly reduce the memory consumptions by 30x for both the network parameters and node attributes, and accelerate the inference by about 47x.

\subsubsection{GNN with Meta Learning}
Recently, several meta learning methods to train GNNs have been proposed to solve the limited samples problem~\cite{mandal2021meta}. Most of the existing works~\cite{huang2020graph,wang2020graph,wen2021meta,ma2020adaptive,jiang2021structure,buffelli2020meta} adopt the Model-Agnostic Meta-Learning algorithm~\cite{finn2017model}. Its outer loop updates the shared parameter, whereas its inner loop updates the task-specific parameter for the current task.
G-META~\cite{huang2020graph} uses local subgraphs to transfer subgraph-specific information and learn transferable knowledge faster via meta gradients with only a handful of nodes or edges in the new task.
AMM-GNN~\cite{wang2020graph} proposes a graph meta-learning framework for attributed graphs, which leverages an attribute-level attention mechanism to capture the distinct information of each task and thus learns more effective transferable knowledge for meta-learning.
MI-GNN~\cite{wen2021meta} studies the problem of inductive node classification across graphs and proposes a meta-inductive framework to customize the inductive model to each graph under a meta-learning paradigm.

\subsection{Edge-Cloud Reinforcement Learning}
Reinforcement Learning (RL) is a special machine learning manner by learning a policy through the interaction between agent and environment, which is close to the human learning way. RL could provide special functionalities such as train-and-error and long-term optimization which is often neglected by traditional unsupervised and supervised learning methods \cite{li2018overviewDRL}. In classical RL studies, it is generally assumed that the agent is an edge device itself or governed by a cloud-based policy \cite{mouradian2018RAAS}. On the other hand, an RL system trained and interacted with the cloud-edge architecture has only been discussed in a limited scope. Nevertheless, there are some thorough investigations for some special domains, which are summarized as follows.

\subsubsection{Federated Reinforcement Learning} 
\begin{figure}[t!]
    \centering
    \includegraphics[width=0.49\textwidth]{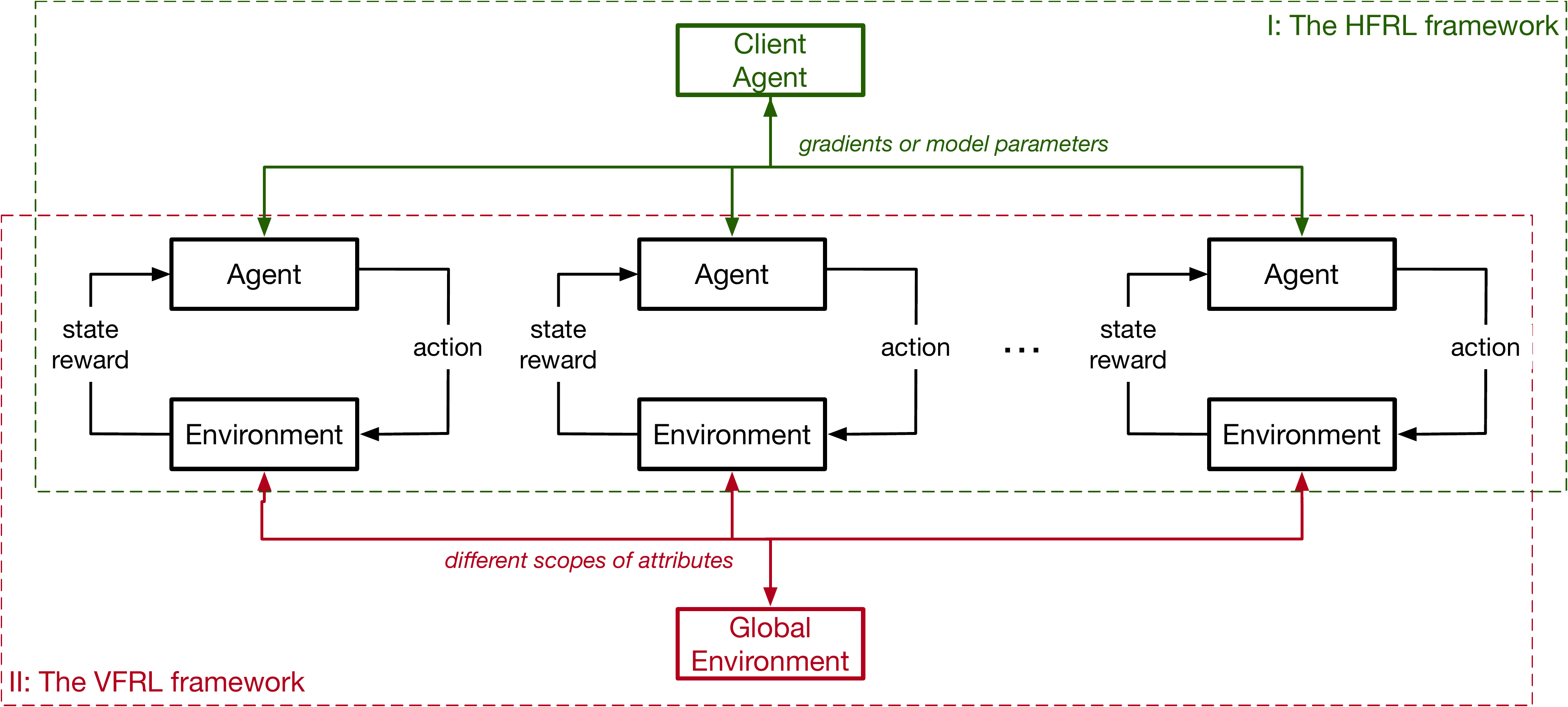}
    \caption{The Federated RL framework. I: the horizontal FRL framework. II: the vertical FRL framework. Each agent-environment inner loop can be deployed on edge, and the client agent or global environment could be cloud-based, making the entire system as edge-cloud collaborative.}
    \label{fig:FRL}
\end{figure}

There are some works which investigate FL~\cite{yang2019fml} within the RL scope~\cite{Zhuo2019FederatedRL, wang2020noniid, yu2021iUDEC, fan2021FegPG-BR, liu2019LFRL}. As summarized in~\cite{qi2021federated}, the idea of FRL could be divided into two main categories: horizontal FRL (HFRL) and vertical FRL (VFRL). Figure~\ref{fig:FRL} illustrates the general frameworks, including HFRL with the server-client architecture, and VFRL with each edge-based environment as a part of the global environment. 

In HFRL, the mobile devices distribute geographically but face similar tasks. The well-studied topic distributed RL is close to HFRL, and HFRL can be viewed as a security-enhanced distributed RL. For example, \cite{wang2020noniid} studies the non-i.i.d. issue of data by deliberately choosing reasonable participating edges during model update. \cite{yu2021iUDEC} proposes a two-timescale (fast and slow) DQN framework and FL is adopted in the fast timescale level and is trained in a distributed manner. \cite{fan2021FegPG-BR} proposes an effective fault tolerance framework given the existence of failed workers. In VFRL, edge devices belong to the same group but their feature spaces are different. VFRL is relatively less studied than HFRL by so far; FedRL~\cite{Zhuo2019FederatedRL} is an example which builds a shared value network and uses the Gaussian differential on information shared by edges. Besides that, Multi-agent RL shares a lot similarities with VFRL; yet VFRL requires modeling on the partially observable Markov decision process while MARL usually assume full observability of system. 

There are also other FRL studies which do not belong to either HFRL or VFRL. For example, \cite{hu2021frs} proposes Federated Reward Shaping in which reward shaping is employed to share federated information of agents. The Multi-task FRL (MT-FedRL)~\cite{anwar2021mtFRL} achieves federation of the policy gradient RL among agents by smoothing average weight. Both works are built on the server-client architecture.

\subsubsection{RL-Assisted Optimization} 
There are also substantial studies which use RL as an aside system in cooperation with mobile-cloud system, to deal with optimization issues including online resource allocation \cite{wu2021disa}, task scheduling~\cite{sheng2021reinforceschedu}, computation offloading~\cite{zhao2020cap, chen2018decentralized, dai2019icct5g, qu2020dmroa, hao2020murl, zhan2020vecppo, wang2018carltm}, and service mitigation \cite{park2020rlservmig}. Among these works, applications are implemented on the Internet of Things~\cite{zhao2020cap, wu2021disa, qu2020dmroa, sheng2021reinforceschedu}, 5G network~\cite{dai2019icct5g}, telemonitoring~\cite{wang2018carltm}, or vehicular terminals~\cite{zhan2020vecppo}. The basic motivation is that those optimization problems are generally NP-hard and easier to solve by DRL built on MDP~\cite{hao2020murl, qu2020dmroa}. The detailed RL methods applied in these attempts include Q-Learning~\cite{wang2018carltm}, DQN~\cite{zhao2020cap, wu2021disa}, REINFORCE~\cite{sheng2021reinforceschedu}, DDPG~\cite{chen2018decentralized}, PPO~\cite{zhan2020vecppo}, and meta-RL~\cite{qu2020dmroa}. Some key elements, such as storage space~\cite{hao2020murl} or context impact~\cite{wang2018carltm}, might also be took into account and \respond{are} used to determined the computation stage of service .

\section{Future Directions and Conclusion}

\subsection{Challenges}
Although AI is widely coming into our life and evolving~\cite{engineering_empower,engineering_airep,engineering_dark}, there are several main challenges for a better Edge-Cloud collaborative AI hindering its development, which we summarize as follows.

\subsubsection{Data}
To our best knowledge, the edge-cloud collaboration open-source datasets are very scarce, which limits the explore in the academic area. The only dataset\footnote{https://tianchi.aliyun.com/dataset/dataDetail?dataId=109858} is released for the Mobile Edge Intelligence in recommendation. The reasons for the real-world open-source datasets are two-fold. On one hand, the fine-grained features on the edge side like the user real-time in-app scrolling, are usually not transmitted to the cloud due to the communication cost and the instant serving bottleneck. That is to say, we cannot completely understand the characteristics of data on the edge side by only training the model on the current cloud-based datasets. On the other hand, for the interactive scenarios like Luoxi~\cite{chen2021mc}, it requires the specific data collection to guarantee the distribution consistency between the training and the test. Simply simulation on the cloud-based dataset by decomposition cannot recover the real-world scenarios. Therefore, more open-source benchmark datasets contributed to this area will promote the research and industrial development. 

\subsubsection{Platform}
The software platform is critical to explore the edge-cloud collaboration, since it is expensive to construct a collaboration environment and simulate the heterogeneous edges and the communication noise. However, it is still in its vacancy to build the well-established platforms that are friendly to a range of algorithmic study. For example, in FL, we have to handle the uncontrollable number of local models uploaded from the edge devices in the real-world scenarios, which may affect the convergence of the training~\cite{kairouz2019advances}. Besides, a systematic analysis on the model training, deployment and evaluation are still lack, which is critical to measure the methods. The current open platforms mainly considering the algorithmic implementation like Luoxi\footnote{https://github.com/luoxi-model/luoxi\_models}. In the future, it will be quite useful to establish the fully functional platform for the open comparison from both academia and industry.

\subsection{Applications}
\subsubsection{Recommendation}

Recommendation systems enable users to find and explore information easily and become increasingly important in a wide range of online applications such as e-commerce, micro-video portals, and social media sites. Despite the huge success, modern recommender systems still suffer from the user-oriented bias/fairness issue~\cite{Li_Chen_Fu_Ge_Zhang_2021,Chen_Dong_Wang_Feng_Wang_He_2020}, privacy leakage~\cite{Shin_Kim_Shin_Xiao_2018,muhammad2020fedfast} and high-latency response~\cite{Freno_Saveski_Jenatton_Archambeau_2015}. For example, centralized on-cloud training will be inevitably biased towards some privileged users, such as active users, resulting in an enlarged performance gap or unfairness among users. Embracing edge-cloud collaborative learning~\cite{dccl} opens up possibilities to address these issues. Expressly, edge training (cloud $\rightarrow$ edge) permits personalization that is free from biased collaborative filtering. Edge models can fully leverage edge features and provide low-latency services, such as dynamic interest modeling, and re-rank. As a counterpart, upon edge models, on-cloud training (edge $\rightarrow$ cloud) benefits from privileged distillation without access to sensitive user data, achieving both privacy protection and full personalization.

\subsubsection{Auto-driving}
Essentially, a vehicle entirely controlled by machines without any human input will possess the acclaimed banner of being autonomous. Nowadays, an important aspect of self-driving vehicles integrating with the cloud is the capability of using the over-the-air electronic communications \cite{Khatun_Gla_Jung_2021}. As well as various sensors for detecting the outside world, the vehicle will be equipped with on-board computer processors and electronic-oriented memory technology, and can communicate with the cloud via a communication device \cite{Vaidya_Kaur_Mouftah_2021}. Besides data communication, edge-cloud collaborative learning techniques will further enhance autonomous driving on safety, functionality, and privacy \cite{Deng_Zhang_Lou_Zheng_Jin_Han_2021}. Initially trained on clouds, edge machine learning models (cloud $\rightarrow$ edge) can interpret real-time raw data, make decisions based on the derived insights, and learn from the feedback from real-time road conditions. Real-time model adaptation is essential for safety and efficiency improvement, and accidents and traffic congestion reduction. Raw data like driving records might contain sensitive contents. Edge models can hereby enhance the centralized training with privacy.

\subsubsection{Games}

Video games have soared as one of the most popular ways to spend time. To ensure a seamless gaming experience, existing games struggle to figure out the work division of cloud computing and edge computing \cite{Nguyen_Tran_Thang_2020,Basiri_Rasoolzadegan_2018,Kassir_Veciana_Wang_Wang_Palacharla_2021}. With edge-cloud collaborative learning, we can leverage the advantages of both sides. On the edge side, we can do more personalization and responsive actions that are sensitive to latency. For example, in sandbox games where the gameplay element is to give players a great degree of creativity and freedom on task completion and Non-Player-Character interaction, personalized AI for the dialogue generation and action-taking are an enticing element. Edge training permits such personalization by taking player behaviors, decisions, and preferences as input. Meanwhile, we can maintain the interactive feedback of edges and the cloud during both training and inference by transferring latent representations or leaving highly intensive but latency-insensitive training tasks for the cloud. Such a bidirectional collaboration brings more creation and freedom to users, ensures a seamless user experience, and opens up new possibilities for advanced gaming systems, such as the Metaverse \cite{Dionisio_Burns_Gilbert_2013,Lee_Braud_Zhou_Wang_Xu_Lin_Kumar_Bermejo_Hui_2021}.

\subsubsection{IoT Security}
Internet of Things security refers to the protection of edges and networks from malicious attacks \cite{Waheed_He_Ikram_Usman_Hashmi_Usman_2021}. Most existing IoT systems are vulnerable to attacks~\cite{Shahid_2021}, where threats in IoT include (but are not limited to) lack of proper data encryption and malicious software. \respond{Currently, some promising explorations~\cite{blockchain1,blockchain2} have been developed for the secure resource management in the blockchain networks, which is an important future direction in the context of edge/cloud computing. However, in a larger landscape, it still requires more consideration about the safety trade-off in the edge-cloud collaboration.} On the one hand, collaborative learning transfers model parameters, latent representations, and back-propagated gradients between edge-edge and edge-cloud, avoiding the communication attack on transferred sensitive data. On the other hand, in bidirectional collaboration, a trustable cloud model can identify malicious edges and mitigate their negative effects, which can be computationally intensive and typically not affordable on edges. 

\subsection{Conclusion}
The rapid development of computing power drives AI to flourish, and evolve into three paradigms: cloud AI, edge AI and edge-cloud collaborative AI. To comprehensively understand the underlying polarization and collaboration of various paradigms, we systematically review the advancement of each direction and build a complete scope. Specially, our survey covers a broad areas including CV, NLP and web services powered by cloud computing, and simultaneously discusses the architecture design and compression techniques that are critical to edge AI. More importantly, we point out the potential collaboration types ranging from privacy-primary collaboration such as federated learning to efficient-primary collaboration for personalization.
We rethink some classical paradigms in the perspective of collaboration that might be extended into edge-cloud collaboration. Some ongoing advanced topics for edge-cloud collaboration are also covered. We summarize the milestone products of cloud computing and edge computing in the recent years, and present future challenges and applications.

\section{Acknowledgement}
This project is partially supported by Natural Science Foundation of China No. U20A20222, the Starry Night Science Fund of Zhejiang University Shanghai Institute for Advanced Study No. SN-ZJU-SIAS-0010 and Young Elite Scientists Sponsorship Program by CAST No. 2021QNRC001.

\ifCLASSOPTIONcaptionsoff
  \newpage
\fi



\balance
\bibliographystyle{IEEEtran}
\bibliography{ref.bib}
\end{document}